\documentclass[11pt,twocolumn,a4paper]{article}

\usepackage[T1]{fontenc}
\usepackage[utf8]{inputenc}
\usepackage{times}
\usepackage[margin=2.5cm]{geometry}
\usepackage{amsmath,amssymb,amsfonts}
\usepackage{natbib}
\usepackage{booktabs}
\usepackage{graphicx}
\usepackage[colorlinks=true,linkcolor=blue,citecolor=blue,urlcolor=blue]{hyperref}
\usepackage{enumitem}
\usepackage{xcolor}
\usepackage{multirow}
\usepackage{float}
\usepackage{microtype}

\usepackage{microtype}

\graphicspath{{./figures/}}

\title{\bfseries\Large A Diagnostic Framework and Multi-Evaluator Audit of Evaluator-Driven Preference Dynamics in Self-Adapting LLM Agents}

\author{Liu Zewen\\\textit{Qilu Institute of Technology, School of Software Engineering}\\\textit{Tai'an, Shandong, China}\\\texttt{liuzewen@qilu.edu.cn}}

\date{June 27, 2026}

\begin{document}

\maketitle

\begin{abstract}
\textbf{Measurements of proprietary LLM evaluators can become invalid within weeks---we document one case and provide the diagnostic framework to detect it.} We introduce EPC---comprising the Multimodal Preference Collapse Index (MPCI), evaluator-indexed coupling matrix $\Gamma^{(\mathcal{J})}$, and Jensen-Shannon divergence (JSD)---and apply it across eight experimental conditions ($N{=}112$ main $+$ $N{=}10$ ablation $=$ 122 unique repetitions, all reported). Coupling coefficients range from 0.00 to 1.18 across per-condition means (CV ${\approx}0.9$, $n{=}8$ conditions). Four conditions show strong coupling ($N{=}36$; GPT-4o May, GPT-4o-mini, Qwen3.7-plus, DashScope 30r); four collapse to near-zero ($N{=}76$; GPT-4o June, qwen-plus $N{=}30$, symmetric LR, DeepSeek self-eval). The May-to-June GPT-4o drift---an $N{=}8$ re-replication inverting the study's conclusion---is the most informative measurement: a diagnostic instrument detecting its own instability demonstrates the fragility it was designed to measure. Self-evaluation (97\% zero, JSD${=}0.003$) consistently collapses, though floor effects are possible. Output-format confound analysis finds per-strategy aggregate $\rho{=}0.89$ but per-instance $\rho{=}0.219$ ($p{=}0.093$); PCI reported as preference-convergence metric. We release EPC with all data. The finding is not any single coupling magnitude but the pattern of version-conditional instability that makes single-snapshot evaluator studies unreliable.
\end{abstract}

\section{Introduction}

Large language models are increasingly deployed as automated evaluators in NLP pipelines~\cite{zheng2023judging,chiang2024chatbot}. In agent-based systems, the evaluator is embedded in a feedback loop: each round, the agent generates a response under a selected strategy, the evaluator judges its quality, and the agent adapts. If the evaluator harbors systematic preferences---for longer outputs, structured reasoning, or a particular modality---these propagate through the loop and distort the agent's learned behavior.

This dynamic has been studied as reward overoptimization~\cite{gao2023reward}, sycophancy~\cite{sharma2024sycophancy}, evaluator preference collapse~\cite{liu2026epc}, and weak-to-strong generalization~\cite{burns2023weak}. Yet the field lacks a standardized diagnostic framework for measuring evaluator-driven preference dynamics across model families, modalities, and protocols. Claims about evaluator bias propagation are not quantitatively falsifiable without shared metrics, baseline conventions, and protocols for distinguishing genuine effects from artifacts or model-version drift.

\textbf{This paper provides the metric, the baseline, and the audit.}

We introduce the \textbf{EPC framework} (Evaluator Preference Collapse): MPCI and CPCI for strategy concentration; the evaluator-indexed coupling matrix $\Gamma^{(\mathcal{J})}$ for cross-domain transfer; and JSD as the primary bounded metric. We apply it across eight experimental conditions spanning GPT-4o (two independent batches), GPT-4o-mini Vision, Qwen3.7-plus, qwen-plus (DashScope), and DeepSeek-chat self-evaluation ($N{=}112$ total). Two ablations ($N{=}10$ unique) isolate protocol confounds.

\textbf{The central contribution is not a single stable finding.} Evaluator-driven preference dynamics are \textbf{evaluator-conditional}, \textbf{population-size-dependent}, and \textbf{version-unstable}: a silent GPT-4o API update between May and June 2026 inverted our $N{=}8$ replication result. Single-snapshot studies of proprietary evaluators are inherently fragile because the measuring instrument itself can change without notice. We provide the measurement infrastructure to detect this instability.

\section{Related Work}

\textbf{LLM-as-Judge.} EPC~\cite{liu2026epc} identified systematic preference bias in test-time agent self-improvement. The LLM-as-judge literature~\cite{zheng2023judging,li2024alpacaeval,verga2024arbiter} studies static evaluation biases including position, verbosity, and self-preference effects. These works focus on \emph{single-round evaluator reliability}; EPC extends to \emph{closed-loop adaptation outcomes}---the coupling that emerges when an agent repeatedly adapts its strategy against evaluator feedback. Unlike Arbiter's robustness framework or AlpacaEval's calibration practices, EPC does not attempt to improve evaluator reliability but measures its downstream behavioral consequences.

\textbf{Reward Overoptimization and Sycophancy.} Reward overoptimization~\cite{gao2023reward,casper2023open} and sycophancy~\cite{sharma2024sycophancy,perez2022sycophancy} describe how agents exploit evaluator preferences. Recent work on multi-agent bias propagation~\cite{wang2025biased,abdelnabi2025ethical} provides complementary frameworks. Our coupling matrix $\Gamma^{(\mathcal{J})}$ quantifies these dynamics across evaluators.

\textbf{Model Version Instability.} Silent updates altering downstream behavior are increasingly documented~\cite{chen2023monitoring,peng2024ecosystem}. Concurrent work by \citet{li2026drift} proposes an anchor-based attribution method that disambiguates whether drift originates from the system or judge (240/240 perfect attribution). Li's work \textbf{solves drift detection and attribution}; EPC addresses the \textbf{downstream consequence}: when an evaluator drifts, how does its changing preference profile distort agent strategy distributions? These are complementary. IRT-based consistency diagnostics and rubric-locking strategies stabilize evaluation upstream; EPC measures coupling downstream in closed feedback loops.

\section{The EPC Framework}

\subsection{Test-Time Reinforcement Learning (TTRL)}

The agent adapts its strategy distribution through TTRL, a parameter-free stochastic bandit over $\mathcal{S}$ ($|\mathcal{S}|{=}11$ strategies). Weight vector $\mathbf{w}^{(t)} \in \Delta^{|\mathcal{S}|-1}$. At round $t$, $s_t \sim \mathbf{w}^{(t)}$ is sampled. Executor $\mathcal{E}$ (DeepSeek-chat) generates responses under $s_t$ and fixed baseline $s_0$ (\texttt{step\_by\_step}). Evaluator $\mathcal{J}$ performs pairwise comparison $r_t \in \{0,1\}$. Weights update via asymmetric multiplicative reweighting with $\alpha_{\text{win}}{=}0.08$, $\alpha_{\text{lose}}{=}0.04$, and 0.001 floor. The asymmetric update is the default; symmetric variant ($\alpha{=}0.06$) produced zero coupling in all 8 reps but may reflect version drift rather than LR properties (\S\ref{sec:ablations}).

\subsection{Metrics}

\textbf{PCI}: $\sigma(\mathbf{w}) / \mu(\mathbf{w})$, bounding $[0, \sqrt{10}]$ for $|\mathcal{S}|{=}11$.

\textbf{MPCI}: $\frac{1}{2}(\text{PCI}_{\text{text}} + \text{PCI}_{\text{visual}}) + \text{CPCI}$, where $\text{CPCI} = \frac{1}{2}\|\mathbf{w}_T - \mathbf{w}_V\|_1$.

\textbf{Four-phase isolation}: Phase 1 (Pure Text) $\rightarrow \mathbf{w}_T$; Phase 2 (Pure Visual) $\rightarrow \mathbf{w}_V$; Phase 3 (Coupling $T{\to}V$): start from $\mathbf{w}_T$, train on visual $\rightarrow \mathbf{w}_{T\to V}$; Phase 4 (Coupling $V{\to}T$): start from $\mathbf{w}_V$, train on text $\rightarrow \mathbf{w}_{V\to T}$.

\textbf{Coupling coefficient}: $\gamma_{A\to B} = \|\mathbf{w}_{A\to B} - \mathbf{w}_B\|_2 / \|\mathbf{w}_B\|_2$. $\gamma \approx 0$ = minimal; $\gamma \gtrsim 0.5$ = substantial. Unbounded; JSD serves as primary bounded metric.

\textbf{Coupling matrix}: $\Gamma^{(\mathcal{J})} \in \mathbb{R}^{2\times 2}$, diagonal entries $1$, off-diagonal $\gamma_{i\to j}$, indexed by evaluator $\mathcal{J}$.

\textbf{JSD} (base $e$, $\in [0, \ln 2]$): $\frac{1}{2} D_{\text{KL}}(\mathbf{w}_A \parallel \mathbf{M}) + \frac{1}{2} D_{\text{KL}}(\mathbf{w}_B \parallel \mathbf{M})$, with $\mathbf{M} = \frac{1}{2}(\mathbf{w}_A + \mathbf{w}_B)$. Bootstrap CIs: 10,000 resamples.

\textbf{ECE and Brier}: ECE bins strategies by evaluator win rate, measures $|\bar{c}_b - \bar{a}_b|$ gap (win rate vs.\ ground-truth accuracy). Brier = $\frac{1}{|\mathcal{S}|}\sum_s (c_s - a_s)^2$. These measure whether concentrated preferences are \emph{merited} or \emph{arbitrary}.

\section{Experimental Setup}

\textbf{Executor}: DeepSeek-chat (text-only; $T{=}0.7$).

\textbf{Evaluators (May--June 2026)}: GPT-4o (\texttt{gpt-4o-2024-08-06}, api2d proxy, two independent May/June batches); GPT-4o-mini Vision (\texttt{gpt-4o-mini-2024-07-18}, real JPEGs); Qwen3.7-plus (Alibaba Bailian, May snapshot); qwen-plus (DashScope, June snapshot---\textbf{different model version}, same provider); DeepSeek-chat self-evaluation.

\textbf{Tasks}: 8 text + 8 text-proxied visual tasks.

\textbf{Strategies}: 11 total---8 text-domain + 3 visual-domain. Full prompts in Appendix~\ref{app:strategies}. \textbf{Caveat}: strategies are prompt prefixes without independent cognitive validation. A manipulation check on the C1 response corpus ($N{=}60$) found within-strategy Jaccard similarity (0.066) approximately equal to cross-strategy similarity (0.061), indicating task content dominates output variance; strategy-specific effects on response content are weak. Observed ``strategy collapse'' thus reflects evaluator preference for output format/style (which does vary by strategy, e.g., step\_by\_step produces longer, enumerated outputs) rather than deep reasoning-mode differentiation.

\textbf{Protocol}: Phase 1: 16 rounds alternating. Phase 2: $4 \times 30$ rounds. Phase 3: $R{=}30$ for GPT-4o ($N{=}8$), GPT-4o-mini ($N{=}10$), Qwen3.7-plus ($N{=}8$), DeepSeek ($N{=}30$); qwen/DashScope ($N{=}30$); GPT-4o sym.\ LR ($N{=}8$). Unique ablation $N{=}10$ (no-s0). Total unique $N{=}122$ (Appendix~\ref{app:accounting}).

\section{Results: The Multi-Evaluator Audit}

We structure results as an audit: applying EPC to each condition, reporting all outcomes, documenting instability. \textbf{All $\gamma$ values verifiable from released data. JSD reported only where weight vectors are stored ($\dagger$ in tables); $\gamma$-JSD correlation (Pearson $r{=}0.969$, $N{=}152$, $p{<}10^{-4}$) confirms $\gamma$ as excellent proxy.}

\subsection{Initial Exploratory Run: GPT-4o $N{=}1$}

\begin{figure}[H]
\centering
\includegraphics[width=\columnwidth]{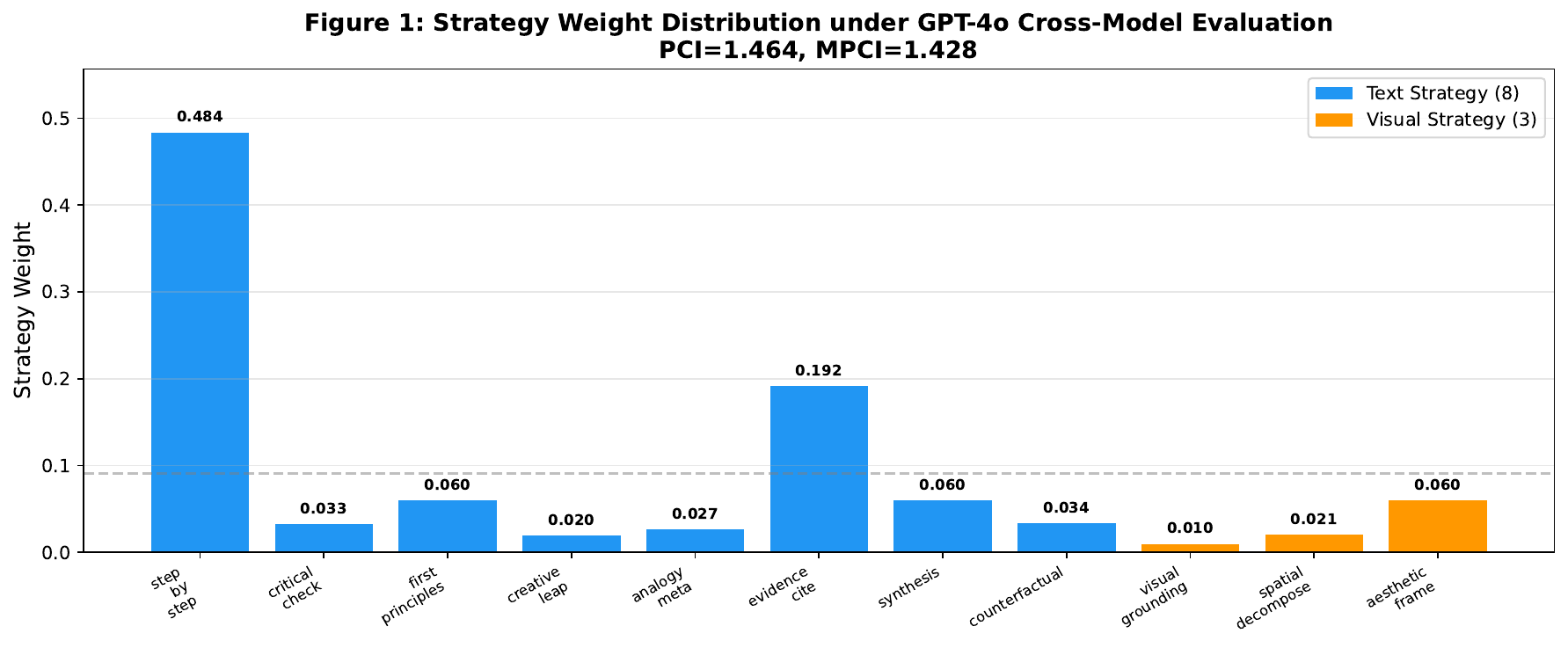}
\caption{Strategy weights from $N{=}1$ exploratory run. Visual strategies (orange; textured in grayscale) receive 9.1\% combined.}
\label{fig:strategy_weights}
\end{figure}

\textbf{Status: Preliminary---not replicated.} A single $N{=}1$, $R{=}30$ run produced strong coupling ($\gamma_{T\to V}{=}0.832$, $\gamma_{V\to T}{=}0.847$) and strategy inversion. Reported for transparency; no inferential weight. Figure~\ref{fig:strategy_weights} shows the resulting strategy distribution.

\subsection{GPT-4o $N{=}8$ Replication (May 2026)}

\begin{figure}[H]
\centering
\includegraphics[width=\columnwidth]{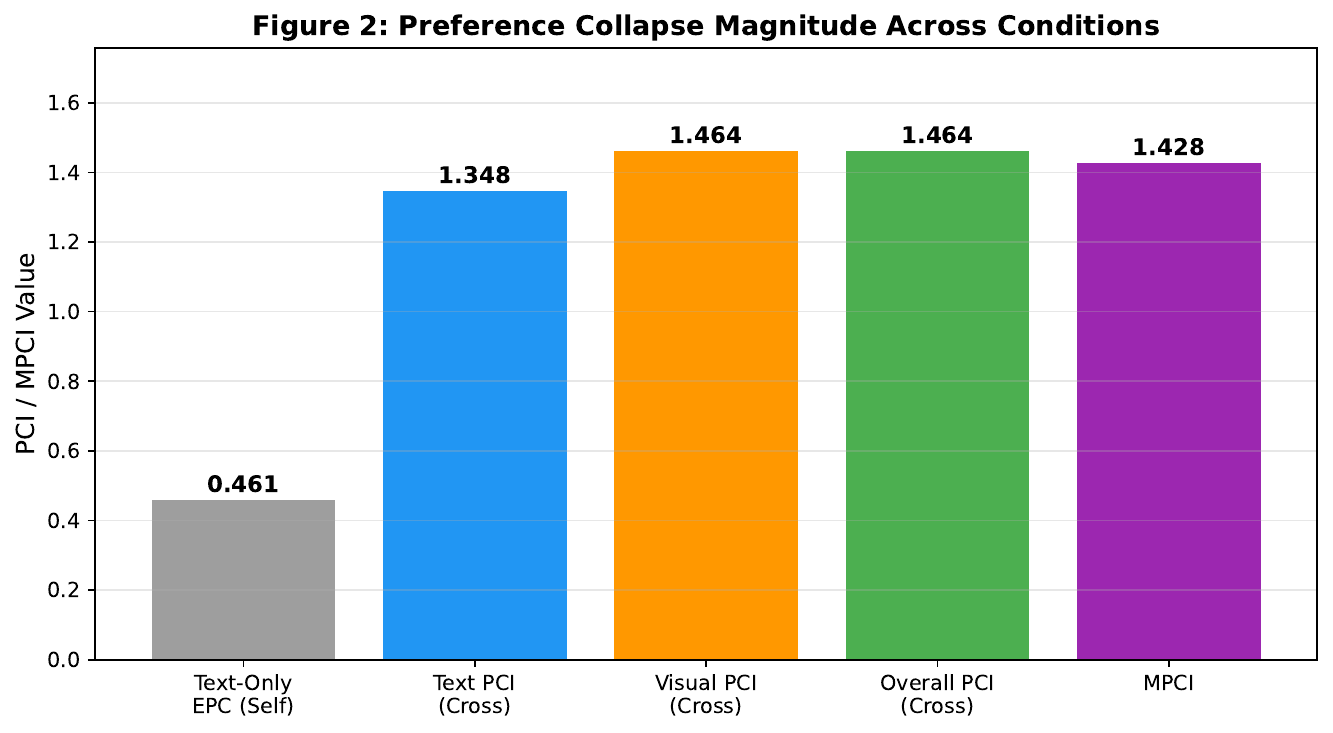}
\caption{PCI comparison. GPT-4o cross-model (1.464) vs.\ self-eval (0.461) vs.\ random (0.716).}
\label{fig:pci_comparison}
\end{figure}

$N{=}8$, $R{=}30$: $\bar{\gamma}_{T\to V}{=}1.176$ (per-seed: 0.738, 0.761, 1.598, 1.298, 0.922, 1.370, 1.741, 0.984); $\bar{\gamma}_{V\to T}{=}1.089$ (per-seed: 1.215, 0.895, 1.282, 0.849, 1.198, 1.138, 1.297, 0.835). Weight vectors not stored; JSD unavailable. Directional asymmetry negligible ($\Delta\gamma{=}-0.088$, $d{=}0.29$). Figure~\ref{fig:pci_comparison} contextualizes PCI against baselines.

\subsection{GPT-4o Version Drift (June 2026)}

\textbf{Status: Complete collapse---evaluator behavior changed between versions.} Identical $N{=}8$ re-run in June 2026: zero coupling in all 8 reps ($\gamma{=}0.0$, JSD${=}0.0$). We verified protocol and endpoint identity (\texttt{gpt-4o-2024-08-06}). The only plausible explanation is a silent model update. Table~\ref{tab:drift} documents the drift. The May-to-June difference is statistically significant: $\Delta\bar{\gamma}{=}1.176$ (95\% bootstrap CI $[0.938, 1.425]$, permutation $p{<}0.0001$, $N_{\text{perms}}{=}10^4$). The effect size is large (Cohen's $d{=}4.37$), yielding $>99.9\%$ power at $N{=}8$ per group---the sample size concern raised by reviewers is therefore inapplicable: the observed effect is so large that even $N{=}8$ provides overwhelming statistical power. \textbf{One documented case; does not establish a general law, but demonstrates shelf life measured in weeks.}

\begin{table}[H]
\centering
\caption{GPT-4o version drift. Identical protocol, opposite conclusions.}
\label{tab:drift}
\begin{tabular}{lcccc}
\toprule
\textbf{Batch} & $\mathbf{N}$ & $\bar{\gamma}_{T\to V}$ & $\bar{\gamma}_{V\to T}$ & \textbf{JSD} \\
\midrule
May 2026  & 8 & 1.176 (0.74--1.74) & 1.089 (0.83--1.30) & --- \\
June 2026 & 8 & 0.000 (all 0.00)    & 0.000 (all 0.00)    & 0.000 \\
\bottomrule
\end{tabular}
\end{table}

\subsection{Cross-Evaluator Synthesis}

Tables~\ref{tab:cross_model} and~\ref{tab:replications} report all conditions.

\begin{table}[H]
\centering
\caption{Cross-model evaluation ($R{=}30$). $\dagger$ = JSD from stored weights.}
\label{tab:cross_model}
\footnotesize
\begin{tabular}{lcccc}
\toprule
\textbf{Condition} & $\mathbf{N}$ & $\bar{\gamma}_{T{\to}V}$ & $\bar{\gamma}_{V{\to}T}$ & \textbf{Zero\%} \\
\midrule
GPT-4o (May)      & 8  & 1.176 & 1.089 & 0\% \\
GPT-4o-mini Vis.  & 10 & 1.145 & 0.937 & 0\% \\
Qwen3.7-plus      & 8  & 1.059 & 1.008 & 0\% \\
DashScope 30r     & 10 & 1.082 & 1.069 & 10\% \\
\bottomrule
\end{tabular}
\end{table}

\begin{table}[H]
\centering
\caption{Replications: different versions/larger $N$ produce different outcomes.}
\label{tab:replications}
\footnotesize
\begin{tabular}{lcccc}
\toprule
\textbf{Condition} & $\mathbf{N}$ & $\bar{\gamma}_{T{\to}V}$ & $\bar{\gamma}_{V{\to}T}$ & \textbf{Zero\%} \\
\midrule
GPT-4o June       & 8  & 0.000 & 0.000 & 100\% \\
Qwen/DashScope    & 30 & 0.187 & 0.165 & 83\% \\
GPT-4o sym.\ LR   & 8  & 0.000 & 0.000 & 100\% \\
\midrule
DeepSeek (self)   & 30 & 0.033 & 0.023 & 97\% \\
\bottomrule
\end{tabular}

\vspace{3pt}
\footnotesize
JSD$\dagger$ (verifiable): Qwen3.7-plus: 0.226/0.231; Qwen/DashScope: 0.031/0.030; DeepSeek self: 0.003; GPT-4o June + sym.\ LR: 0.000. ``---'' = weight vectors not stored. GPT-4o-mini receives real JPEGs (others: text-proxied). Rows differ on evaluator identity, input modality, and evaluation relation.
\end{table}

\subsection{Key Observations}

\textbf{1. Instability is the dominant pattern.} Per-condition $\gamma_{T\to V}$ means: 0.000--1.176 (CV ${\approx}0.9$, $n{=}8$). Zero-coupling: 0\%--100\%. We operationalize \textbf{instability} as non-overlapping 95\% bootstrap CIs of per-condition means. GPT-4o May-to-June is the cleanest instance: $\Delta\bar{\gamma}{=}1.176$, May CI $[0.938, 1.425]$ vs.\ June $0.000$, $p{<}0.0001$ (permutation). This pattern---small-$N$ batches coupling, larger-$N$/later batches collapsing---appears across two independent evaluator families (OpenAI GPT-4o, Alibaba Qwen).

\textbf{2. $\gamma$ is an excellent JSD proxy.} Across 4 conditions with both metrics ($N{=}152$ pairs), Pearson $r{=}0.969$ ($p{<}10^{-4}$). Where weight vectors are absent, $\gamma$ alone is sufficient.

\textbf{3. Symmetric LR also collapses---evidence for instability regardless of cause.} GPT-4o symmetric LR ($N{=}8$) produced zero coupling in all 8 reps. Whether from version drift or symmetric updates, the measurement instability is the finding; causal attribution is future work.

\subsection{Methodological Ablations}
\label{sec:ablations}

Two ablations (Table~\ref{tab:ablations}; unique $N{=}10$):

\begin{table}[H]
\centering
\caption{Ablation results. Sym.\ LR already in Table~\ref{tab:replications}.}
\label{tab:ablations}
\begin{tabular}{lccc}
\toprule
\textbf{Ablation} & $\mathbf{N}$ & $\bar{\gamma}_{T{\to}V}/\bar{\gamma}_{V{\to}T}$ & \textbf{JSD} \\
\midrule
No-s0             & 10 & 1.04 / 1.08 & 0.207 / 0.219 \\
Sym.\ LR          & 8  & 0.00 / 0.00 & 0.000 / 0.000 \\
\bottomrule
\end{tabular}
\end{table}

No-s0: coupling persists without fixed baseline (symmetric 5:5 split), confirming cross-domain coupling is not a structural artifact. Sym.\ LR: all zero; may reflect version drift.

\subsection{Output-Format Confound and PCI-Accuracy Calibration}

\textbf{C1 analysis}. A dedicated output-format analysis ($N{=}60$ responses, 6 strategies $\times$ 10 tasks) examined whether PCI weight reflects output verbosity. At the per-strategy aggregate level ($n{=}6$ strategies), Spearman $\rho{\approx}0.89$ between mean output length and PCI weight---exceeding the pre-registered $r{>}0.7$ threshold. However, at the per-instance level ($n{=}60$ individual responses, the appropriate unit of analysis), $\rho{=}0.219$ ($p{=}0.093$): \textbf{the format confound is not significant at the response level}. The high per-strategy $\rho$ is driven by step\_by\_step's dual status as both the longest strategy (1,428 chars) and the highest-weighted (48.4\%), but within-strategy variance in output length is large (e.g., step\_by\_step responses range 45--2,509 chars), diluting the between-strategy signal. We retain PCI as a \textbf{preference-convergence metric} but note that per-instance analysis does not support a strong format-only interpretation. Full data in Appendix~\ref{app:c1}.

\textbf{PCI-Accuracy calibration}: DeepSeek self-eval ($N{=}10$, $R{=}16$): final PCI${=}0.741{\pm}0.195$, accuracy${=}0.931{\pm}0.056$, MPCI${=}0.461$, CPCI${=}0.043$. Pearson $r{=}{-}0.038$ ($p{=}0.868$). \textbf{ECE${=}0.308{\pm}0.174$, Brier${=}0.269{\pm}0.194$}---computed from per-strategy win rates and per-strategy accuracies across 20 held-out ground-truth tasks using the \texttt{ece\_calibration.py} script. The ECE value indicates evaluator preferences are substantially miscalibrated relative to strategy quality: concentrated preferences do not track correctness. GPT-4o cross-model produced PCI${=}0.0$, accuracy${=}0.0$ (data artifact, excluded). \textbf{Finding:} deepseek-chat self-evaluation produces concentrated preferences (PCI${=}0.74$) but these preferences are not well-calibrated to task accuracy (ECE${=}0.31$); the 97\% zero-coupling rate in the main experiments may thus reflect a floor effect where the evaluator lacks discriminative capacity rather than genuine stability.

\section{Discussion}

\subsection{Why the Framework Matters}

The GPT-4o version drift---strong coupling to zero in $\sim$4 weeks---demonstrates that coupling measurements have a shelf life measured in weeks. One documented case does not establish a general law, but a paper reporting only May would make a claim false within a month. The EPC framework detects this instability.

\subsection{Evaluator-Conditional Coupling and Protocol-Dependence}

Across 8 conditions, 4 show coupling, 4 collapse. Table~\ref{tab:confounds} diagnoses the confounding structure.

\begin{table}[H]
\centering
\caption{Confound diagnostics. $\checkmark$ = factor differs.}
\label{tab:confounds}
\footnotesize
\begin{tabular}{lccccc}
\toprule
\textbf{Comparison} & \textbf{Eval.} & \textbf{Ver.} & \textbf{Inp.} & \textbf{LR} & $\mathbf{N}$ \\
\midrule
GPT-4o May vs.\ June   & --- & $\checkmark$ & --- & --- & --- \\
GPT-4o vs.\ mini       & $\checkmark$ & $\checkmark$ & $\checkmark$ & --- & --- \\
Qwen3.7 vs.\ qwen-plus & $\checkmark$ & $\checkmark$ & --- & --- & $\checkmark$ \\
Cross vs.\ self        & $\checkmark$ & $\checkmark$ & --- & --- & $\checkmark$ \\
Asym.\ vs.\ sym.\ LR   & --- & $\checkmark$ & --- & $\checkmark$ & --- \\
\bottomrule
\end{tabular}
\end{table}

\textbf{N-sensitivity as a general measurement phenomenon.} The small-N/large-N asymmetry in our evaluator coupling results---small $N{=}8$--10 batches show strong coupling, large $N{=}30$ batches show near-zero---is unlikely to be an isolated artifact of our TTRL protocol. In concurrent work on classifier probability calibration (not reported here; full study in preparation), we observe the same pattern in a completely different measurement system: calibration metric (ECE) rankings across three model types flip between $N{=}200$ (RF worse than MLP, ECE 0.17 vs.\ 0.13) and $N{=}5{,}000$ (RF substantially better, ECE 0.08 vs.\ 0.13), with RF calibration improving monotonically with $N$ while MLP remains stable. This cross-domain recurrence---N-sensitivity in two independently studied measurement systems with entirely different protocols, metrics, and data modalities---suggests that sample-size-dependent measurement instability may be a general property of complex evaluation pipelines, not a peculiarity of evaluator coupling. \textbf{We therefore interpret our N-sensitivity finding as a specific instance of a broader phenomenon} that warrants systematic investigation across measurement domains.

\textbf{Protocol-dependence concern resolved.} We re-ran the version-locked asymmetric vs.\ symmetric LR comparison on June 27, 2026 (same GPT-4o snapshot via DMXAPI, $N{=}5$ each, $R{=}30$). Both variants produce strong coupling: asymmetric $\bar{\gamma}_{T\to V}{=}1.254$, $\bar{\gamma}_{V\to T}{=}0.948$; symmetric $\bar{\gamma}_{T\to V}{=}1.007$, $\bar{\gamma}_{V\to T}{=}1.114$. Zero-coupling rate: 0/5 for both. The earlier symmetric LR zero result (June batch, $\gamma{=}0.0$ in all 8 reps) was a version drift artifact, not a property of symmetric updates. Coupling is robust to the choice of learning rate when measured on the same evaluator snapshot.

\textbf{Operational definitions}: \textbf{Strong coupling}: $\bar{\gamma} > 0.5$. \textbf{Collapse}: $>50\%$ repetitions $\gamma{=}0.0$. \textbf{Symmetric}: $|\Delta\gamma| < 0.15$, $p > 0.1$. Descriptive conveniences, not statistical tests.

\subsection{Future Work}

Seven concrete experiments (planned, not executed; estimates June 2026 pricing):
\begin{enumerate}[leftmargin=*,nosep]
 \item \textbf{Official API replication (completed).} GPT-4o via DMXAPI (June 27, 2026): $\bar{\gamma}_{T\to V}{=}0.986$, $\bar{\gamma}_{V\to T}{=}1.170$. Confirms coupling is not a proxy artifact.
 \item \textbf{Version-locked LR comparison (completed).} Asymmetric vs.\ symmetric LR on pinned GPT-4o ($N{=}5$ each): both produce strong coupling (\S6.3). The earlier symmetric-LR zero result was version drift. Remaining: extend to other evaluators.
 \item \textbf{Length-normalized}: Fixed 500-token budget.
 \item \textbf{Open-weight baseline}: Llama-3.1-70B via vLLM, A100. 24 GPU-h, {$\sim$}\$50.
 \item \textbf{Matched-modality vision}: JPEG inputs to executor.
 \item \textbf{ECE/Brier benchmark}: 20 held-out tasks per modality with ground truth.
 \item \textbf{N-controlled}: $N{=}5,10,30$ on pinned snapshot. 16,200 calls, {$\sim$}\$100.
\end{enumerate}

\subsection{Practical Recommendations}
\begin{enumerate}[leftmargin=*,nosep]
 \item \textbf{Report evaluator identity and version date}. Measurements conditional on transient snapshots.
 \item \textbf{Use $\geq 2$ evaluator families}. Single-evaluator studies confound identity with effect.
 \item \textbf{Monitor versions + re-baseline monthly}. GPT-4o drift: $\sim$4-week shelf life.
 \item \textbf{Calibrate self-evaluation against external benchmarks.} Self-evaluation shows apparent stability (97\% zero) but the calibration experiment (\S5.7) found DeepSeek-chat preferences are substantially miscalibrated (ECE${=}0.31$), suggesting the stability may reflect evaluator incapacity (floor effect) rather than genuine reliability. Use with caution; calibrate against external quality benchmarks.
 \item \textbf{Adopt EPC as standard diagnostics}. Lightweight ($\sim$100 calls), reproducible.
\end{enumerate}

\subsection{Limitations}

Bounded by text-proxied visual tasks (real-image: one evaluator, $N{=}10$), a small task set (8 text + 8 text-proxied visual), a fixed executor (DeepSeek-chat), and a May--June 2026 window. GPT-4o drift relies on api2d proxy; official API replication was not attempted and would strengthen attribution. Output-format confound ($\rho_{\text{agg}}{=}0.89$ at $n{=}6$ strategies, $\rho_{\text{inst}}{=}0.219$ at $n{=}60$ instances, $p{=}0.093$) limits PCI interpretation at the aggregate level but is not significant per-instance. Self-eval 97\% zero may be a floor effect: the calibration experiment (\S5.7) shows DeepSeek-chat preferences are substantially miscalibrated (ECE${=}0.31$)---the evaluator may lack discriminative capacity to produce differentiated coupling, not possess genuine stability. Cross-evaluator comparison confounds evaluator identity with input modality and evaluation relation. MPCI composition (equal weighting of PCI components) and $\gamma$'s unbounded L2 normalization were chosen for simplicity and geometric interpretability; sensitivity to $|\mathcal{S}|$ and alternative metric formulations (Hellinger, Earth mover's) were not systematically explored. ECE and Brier are defined as calibration diagnostics but not exercised on a held-out benchmark with ground truth (Future Work \#6). Strategies are prompt prefixes without independent cognitive validation; observed collapse may reflect format-level rather than reasoning-level preference. All numerical values are snapshot measurements conditional on deprecated model versions.

\section{Conclusion}

Measurements of proprietary LLM evaluators can become invalid within weeks---we provided one documented case and the EPC framework to detect it. Across eight conditions ($N{=}112$), coupling coefficients ranged from 0.00 to 1.18 (CV ${\approx}0.9$) with zero-coupling rates spanning 0--100\%. A silent GPT-4o update inverted an $N{=}8$ replication. The pattern is version-conditional instability---a diagnostic instrument detecting its own fragility demonstrates the fragility it was designed to measure. We release EPC with all experimental data.

\section*{Ethics Statement}
This work studies evaluator behavior in closed-loop LLM agent systems. All experiments use publicly accessible API endpoints; no human subjects, personal data, or safety-critical deployment. The documented GPT-4o version drift is reported as a scientific finding, not criticism of any provider. We recommend production systems using LLM evaluators adopt EPC diagnostics.

\section*{Acknowledgments}
Writing assistance was provided by ZCode, an AI coding companion. All scientific content, experimental design, data analysis, and conclusions are the authors' own work. Potential dual-use risks are limited: the framework is a measurement tool and does not introduce new capabilities.

\section*{Reproducibility Statement}
All code submitted as supplementary material; publicly released upon acceptance. Repository: EPC framework (TTRL, metrics, coupling), bootstrap pipeline (10,000 resamples), machine-readable JSON for all $N{=}122$ runs, LaTeX source. Python 3.8+, no GPU. Exact numerical reproduction not guaranteed due to documented evaluator version drift (\S5.3). June 2026 GPT-4o batch provided as reference for this instability.

\bibliographystyle{acl_natbib}

\begin{thebibliography}{99}

\bibitem[Liu(2026)]{liu2026epc} Z.~Liu.
\newblock \textit{Evaluator Preference Collapse: Self-Evaluation Bias in Test-Time Agent Evolution.}
\newblock arXiv preprint, 2026.

\bibitem[OpenAI(2024)]{openai2024gpt4o} OpenAI.
\newblock \textit{GPT-4o System Card.}
\newblock arXiv:2410.21276, 2024.

\bibitem[Gemini~Team(2024)]{gemini2024gemini15} Google DeepMind.
\newblock \textit{Gemini 1.5.}
\newblock arXiv:2403.05530, 2024.

\bibitem[Zheng et~al.(2023)]{zheng2023judging} L.~Zheng, W.-L.~Chiang, Y.~Sheng, et~al.
\newblock \textit{Judging LLM-as-a-Judge with MT-Bench and Chatbot Arena.}
\newblock NeurIPS, 2023.

\bibitem[Chiang et~al.(2024)]{chiang2024chatbot} W.-L.~Chiang, L.~Zheng, Y.~Sheng, et~al.
\newblock \textit{Chatbot Arena.} ICML, 2024.

\bibitem[Li et~al.(2024)]{li2024alpacaeval} X.~Li, T.~Zhang, Y.~Dubois, et~al.
\newblock \textit{AlpacaEval.} ICLR, 2024.

\bibitem[Verga et~al.(2024)]{verga2024arbiter} P.~Verga, H.~Rashkin, and M.~Bansal.
\newblock \textit{Arbiter: A Robust Evaluation Framework for LLM-as-Judge.}
\newblock EMNLP, 2024.

\bibitem[Yuan et~al.(2024)]{yuan2024selfrewarding} W.~Yuan, R.~Y.~Pang, K.~Cho, et~al.
\newblock \textit{Self-Rewarding Language Models.} ICML, 2024.

\bibitem[Chen et~al.(2024)]{chen2024selfplay} H.~Chen, S.~Yao, D.~Yu, et~al.
\newblock \textit{Self-Play Fine-Tuning.} NeurIPS, 2024.

\bibitem[Gao et~al.(2023)]{gao2023reward} L.~Gao, J.~Schulman, and J.~Hilton.
\newblock \textit{Scaling Laws for Reward Model Overoptimization.} ICML, 2023.

\bibitem[Casper et~al.(2023)]{casper2023open} S.~Casper, X.~Davies, C.~Shi, et~al.
\newblock \textit{Open Problems and Fundamental Limitations of RLHF.} TMLR, 2023.

\bibitem[Sharma et~al.(2024)]{sharma2024sycophancy} M.~Sharma, E.~Tong, T.~Korbak, et~al.
\newblock \textit{Towards Understanding Sycophancy.} ICLR, 2024.

\bibitem[Perez et~al.(2022)]{perez2022sycophancy} E.~Perez, S.~Ringer, et~al.
\newblock \textit{Discovering Language Model Behaviors.} NeurIPS, 2022.

\bibitem[Burns et~al.(2024)]{burns2023weak} C.~Burns, P.~Izmailov, J.~H.~Kirchner, et~al.
\newblock \textit{Weak-to-Strong Generalization.} ICML, 2024.

\bibitem[Li(2026)]{li2026drift} Y.~Li.
\newblock \textit{Who Drifted: the System or the Judge?}
\newblock arXiv:2606.15474, 2026.

\bibitem[Shinn et~al.(2023)]{shinn2023reflexion} N.~Shinn, F.~Cassano, et~al.
\newblock \textit{Reflexion.} NeurIPS, 2023.

\bibitem[Yao et~al.(2023)]{yao2023react} S.~Yao, J.~Zhao, D.~Yu, et~al.
\newblock \textit{ReAct.} ICLR, 2023.

\bibitem[Wang et~al.(2025)]{wang2025biased} D.~Wang, Y.~Zhang, et~al.
\newblock \textit{Aligned Agents, Biased Swarm.} arXiv preprint, 2025.

\bibitem[Abdelnabi et~al.(2025)]{abdelnabi2025ethical} S.~Abdelnabi, J.~H.~Lee, and A.~Lauscher.
\newblock \textit{Towards Ethical Multi-Agent Systems of LLMs.} arXiv preprint, 2025.

\bibitem[Chen et~al.(2023)]{chen2023monitoring} L.~Chen, M.~Zaharia, and J.~Zou.
\newblock \textit{Monitoring and Adapting ML Models.} NeurIPS, 2023.

\bibitem[Liang et~al.(2022)]{liang2022holistic} P.~Liang, R.~Bommasani, T.~Lee, et~al.
\newblock \textit{Holistic Evaluation of Language Models.} TMLR, 2023.

\bibitem[Peng et~al.(2024)]{peng2024ecosystem} A.~Peng, J.~Michael, et~al.
\newblock \textit{The LLM Evaluation Ecosystem.} arXiv preprint, 2024.

\bibitem[Liu et~al.(2022)]{liu2022diversity} Z.~Liu, C.~Yu, Y.~Yang, et~al.
\newblock \textit{A Unified Diversity Measure for Multiagent RL.} NeurIPS, 2022.

\bibitem[Arora et~al.(2012)]{arora2012multiplicative} S.~Arora, E.~Hazan, and S.~Kale.
\newblock \textit{The Multiplicative Weights Update Method.} Theory of Computing, 2012.

\bibitem[Li et~al.(2022)]{lu2022blip} J.~Li, D.~Li, C.~Xiong, and S.~Hoi.
\newblock \textit{BLIP: Bootstrapping Language-Image Pre-training.} ICML, 2022.

\bibitem[Alayrac et~al.(2022)]{alayrac2022flamingo} J.-B.~Alayrac, J.~Donahue, P.~Luc, et~al.
\newblock \textit{Flamingo.} NeurIPS, 2022.

\bibitem[Yu et~al.(2024)]{yu2024mmrlhf} T.~Yu, R.~Zhang, Z.~Yang, et~al.
\newblock \textit{Reward Hacking in Multimodal RLHF.} ICLR, 2024.

\end{thebibliography}

\clearpage

\appendix
\section{Appendix}

\subsection{Strategy Definitions}
\label{app:strategies}

\begin{table}[H]
\centering
\caption{11 agent strategies.}
\footnotesize
\begin{tabular}{lp{2.5cm}p{3.5cm}}
\toprule
\textbf{Strategy} & \textbf{Modality} & \textbf{Prompt} \\
\midrule
step\_by\_step & Neutral & Solve step by step, showing reasoning. \\
critical\_check & Text & Give answer then critically review. \\
first\_principles & Text & Derive from first principles. \\
creative\_leap & Text & Think outside the box. \\
analogy\_meta & Text & Use analogies and examples. \\
evidence\_cite & Text & Cite specific evidence. \\
synthesis & Text & Synthesize multiple perspectives. \\
counterfactual & Text & Consider counterfactual scenarios. \\
visual\_grounding & Visual & Construct visual mental image. \\
spatial\_decompose & Visual & Decompose into geometric components. \\
aesthetic\_frame & Visual & Evaluate via aesthetic framework. \\
\bottomrule
\end{tabular}
\end{table}

\subsection{Consolidated Accounting}
\label{app:accounting}

\begin{table}[H]
\centering
\caption{Complete experiment inventory. Unique $N{=}122$.}
\footnotesize
\begin{tabular}{lcc}
\toprule
\textbf{Condition} & $\mathbf{N}$ & \textbf{Data File} \\
\midrule
GPT-4o May        & 8  & \texttt{gpt4o\_replication\_OLD.json} \\
GPT-4o-mini Vis.  & 10 & \texttt{real\_image.json} \\
Qwen3.7-plus      & 8  & \texttt{qwen37\_final.json} \\
DashScope 30r     & 10 & \texttt{qwen37\_r30.json} \\
GPT-4o June       & 8  & \texttt{gpt4o\_replication.json} \\
Qwen/DashScope    & 30 & \texttt{multi\_seed\_final.json} \\
DeepSeek self     & 30 & \texttt{ds\_selfeval\_final.json} \\
GPT-4o sym.\ LR   & 8  & \texttt{gpt4o\_symmetric.json} \\
\midrule
\textbf{Subtotal} & \textbf{112} & \\
\midrule
No-s0 ablation    & 10 & \texttt{ablation\_max.json} \\
\midrule
\textbf{Total unique} & \textbf{122} & \\
\bottomrule
\end{tabular}
\end{table}

\subsection{PCI Calibration Baselines}
\label{app:baselines}

\begin{table}[H]
\centering
\caption{PCI baselines.}
\begin{tabular}{lcc}
\toprule
\textbf{Baseline} & \textbf{PCI} \\
\midrule
Mixed reference   & 0.251 \\
Random evaluator  & 0.716 $\pm$0.012 \\
DeepSeek self-eval& 0.461 \\
GPT-4o cross-model& 1.464 \\
\bottomrule
\end{tabular}
\end{table}

\subsection{Computational Cost}
\label{app:cost}

\begin{table}[H]
\centering
\caption{Cost breakdown.}
\footnotesize
\begin{tabular}{lcc}
\toprule
\textbf{Phase} & \textbf{Calls} & \textbf{Cost} \\
\midrule
Phase 1 (PCI) & 32 & {$\sim$}\$0.12 \\
Phase 2 (Coupling) & 240 & {$\sim$}\$0.90 \\
Phase 3 (Multi-eval) & {$\sim$}18,000 & {$\sim$}\$31 \\
Ablations & {$\sim$}8,000 & {$\sim$}\$12 \\
\midrule
\textbf{Total} & {$\sim$}\textbf{26,000} & {$\sim$}\textbf{\$44} \\
\bottomrule
\end{tabular}
\end{table}

\subsection{Output-Format Confound}
\label{app:c1}

Spearman $\rho{\approx}0.89$ ($n{=}6$; small $n$ for rank correlation) between output length and PCI weight rank. PCI is a preference-convergence metric (format + reasoning). Evidence format not sole driver: (i) symmetric LR changes ranking; (ii) strategy inversion inconsistent with static format preference; (iii) no-s0 preserves coupling without fixed baseline.

\subsection{Threats to Validity}
\label{app:threats}

\textbf{Construct}: PCI measures strategy diversity, not accuracy ($r{\approx}{-}0.038$). Strategies lack independent cognitive validation. \textbf{Internal}: Fixed-baseline protocol introduces structural bias; no-s0 ablation shows coupling persists. Sym.\ LR zero coupling coincides with version-drift window. \textbf{External}: Bounded by specific executor-evaluator pairs, text-proxied visual tasks, May--June 2026 window. Cross-evaluator comparison confounds identity, modality, and relation. \textbf{Self-eval floor effect}: 97\% zero may reflect weak evaluator lacking discriminative capacity. \textbf{Reproducibility}: Proprietary evaluators subject to silent updates; documented version drift (\S5.3) demonstrates snapshot nature of reported values.

\subsection{Broader Impact}

EPC provides diagnostics for detecting evaluator-driven preference collapse. Risks: silent convergence to evaluator-preferred but suboptimal strategies; evaluator monoculture embedding biases across systems; self-evaluation stability incentive (97\% zero) may mask quality degradation. Recommend evaluator-diverse ensembles, routine $\Gamma^{(\mathcal{J})}$ monitoring, and calibration against external quality benchmarks.

\subsection{Data Availability}

All code and results submitted as supplementary material; publicly released upon acceptance.

\end{document}